# Detecting Violent and Abnormal Crowd activity using Temporal Analysis of Grey Level Co-occurrence Matrix (GLCM) Based Texture Measures

**Kaelon Lloyd · Paul L. Rosin · David Marshall · Simon C. Moore**

**Abstract** The severity of sustained injury resulting from assault-related violence can be minimised by reducing detection time. However, it has been shown that human operators perform poorly at detecting events found in video footage when presented with simultaneous feeds. We utilise computer vision techniques to develop an automated method of abnormal crowd detection that can aid a human operator in the detection of violent behaviour. We observed that behaviour in city centre environments often occur in crowded areas, resulting in individual actions being occluded by other crowd members. We propose a real-time descriptor that models crowd dynamics by encoding changes in crowd texture using temporal summaries of Grey Level Co-Occurrence Matrix (GLCM) features. We introduce a measure of inter-frame uniformity (IFU) and demonstrate that the appearance of violent behaviour changes in a less uniform manner when compared to other types of crowd behaviour. Our proposed method is computationally cheap and offers real-time description. Evaluating our method using a privately held CCTV dataset and the publicly available Violent Flows, UCF Web Abnormality, and UMN Abnormal Crowd datasets, we report a receiver operating characteristic score of 0.9782, 0.9403, 0.8218 and 0.9956 respectively.

K. Lloyd
School of Computer Science, Cardiff University
E-mail: Lloydk1@cardiff.ac.uk

P.L. Rosin
School of Computer Science, Cardiff University
E-mail: Paul.Rosin@cs.cardiff.ac.uk

D. Marshall
School of Computer Science, Cardiff University
E-mail: MarshallAD@cardiff.ac.uk

S.C. Moore
Violence and Society Research Group, Applied Clinical Research and Public Health, School of Dentistry, Cardiff University
E-mail: mooresc2@cardiff.ac.uk

## 1 Introduction

City centre locations around the world are characterized by the presence of surveillance cameras. One typical use of these cameras is to aid law enforcement by allowing operators to actively identify criminal activity. It is estimated that in the United Kingdom alone there are upwards of 1.8 million Closed Circuit Television (CCTV) cameras installed across both public and private sectors, or about one camera for every 35 people, with the average person falling into the viewshed of a camera system at least 68 times a day [12,27]. An issue with having such a large number of surveillance cameras is that they capture too much data for effective human observation. A study undertaken by Voorthuijsen et al. [32] investigated the human ability to detect scenes of interest from video data when presented with different numbers of simultaneous video feeds. On average, the human ability to detect scenes of interest dropped by 19% when the number of simultaneous feeds was increased from one to four. It is reasonable to assume that the observed drop in event detection ability becomes greater when a single person is subject to the much larger video arrays common in modern observation centres.

Evidence suggests that surveillance systems can reduce the incidence rate of hospital assault-related attendances. Sivarajasingam et al. [31] investigated the relationship between surveillance system installation on violence. It was shown that police recorded violence increased but hospital admissions for assault-related injury fell, an effect that the authors suggest is due to earlier police intervention preventing disorder escalating to the point where serious injury is inflicted on victims of violence. This finding was corroborated by research undertaken by Florence et al. [9] that evaluated data sharing schemes focused on effective strategies for reducing violence. The authors of this study highlight the importance of rapid intervention in reducing injury severity. Although surveillance systems were not the sole focus of this research their usefulness at allowing for effective intervention is acknowledged. A follow up evaluation of the study undertaken by Florence et al. [9] asserts a £1.2 million saving after the application of violence reduction strategies in the year 2007 [10].

There are two practical limitations in using real-world surveillance footage, firstly, the cost of upgrading video capture devices used in surveillance systems is high, and only cameras that are deemed important are upgraded. It is therefore common for video surveillance systems to be composed of both modern and legacy hardware components. The quality of recorded footage from older cameras is typically poor due to hardware limitations and it is common for footage to have low spatial and temporal resolutions. Additionally, outdoor CCTV cameras are subject to natural illumination changes that result in poor contrast when recording footage at night, which can make effective description of content difficult. Second, footage of dense populations in urban environments tends to depict moving, self-occluding crowds. It can be difficult to generate a meaningful description of individuals and their actions as the visual consistency of recognizable shapes fluctuates greatly between frames due to the high levels of occlusion. In recognizing the potential value of surveillance systems in reducing assault-related injury this paper describes a novel solution to the previously discussed limitations, and therefore provides opportunities for the classification of live surveillance feeds to aid operators' early ascertainment of disorderly behaviour and by extension, their capacity to direct resources that stop disorder escalating.

Our novel solution builds upon the idea of describing crowded scenes using visual texture. Texture is well suited for describing the seemingly unstructured patterns that result from the mass occlusions caused by crowding [22]. We assert that the appearance of abnormally behaving crowds undergo different patterns of change when compared to crowds exhibiting normal behaviour. We therefore propose a description based on encoding the change in crowd appearance over time. This is accomplished by computing texture features on a per-frame basis for a sequence of video frames and summarising how the texture features evolve. We present a computationally cheap method of abnormal crowd description that achieves state-of-the-art results across many datasets, including our own real-world CCTV dataset known as CF-Violence. We show that our proposed method generates a scene description that can be used to discriminate between abnormal and normal scenes available in the *UMN unusual crowd*, and *UCF Web abnormality* datasets. We also demonstrate state-of-the-art discrimination between violent and non-violent scenes as shown in the *Violent Flows* and *CF-Violence* datasets. We also provide an extensive investigation of the parameter effects of our proposed method and demonstrate that violent behaviour holds a property of non-uniform change over time, a property that to the authors knowledge, has not been previously discussed in the context of crowd behaviour.

## 2 Related Work

A wide selection of approaches aiming to solve the crowd abnormality detection problem have been proposed over the years. Kratz *et al.* [19] and Marques *et al.* [23] state that tracking features in extremely dense, complex crowds is infeasible and that optical flow approximation can become unreliable. Kratz *et al.* [19] avoid optical flow based motion description by extracting fixed size spatio-temporal volumes and computing spatio-temporal gradients of pixel intensities which are represented using a 3D Gaussian Mixture Model (GMM). The authors model normal behaviour using a Hidden Markov Model and declare a new observation as abnormal if it does not fit the learnt model. Wang *et al.* [34] also avoid an optical flow based representation when dealing with crowds, they instead favour statistics computed from wavelet transformed spatio-temporal slices taken from a spatio-temporal volume. Although the effectiveness of optical flow is often a point of theoretical contention when dealing with crowds, there exists many methods that utilize measures of optical flow with great results. Ryan *et al.* [7] encode optical flow vectors using a 3 dimensional Grey-Level Co-occurrence Matrix (GLCM) structure, expressing the dynamics of a local region by the uniformity of motion. The authors generate a model of normality using a Gaussian Mixture Model. The authors claim that their method is both effective at discerning between normal and abnormal scenes while maintaining an arguably real-time processing speed of approximately 9 frames per second. Wang *et al.* [35] compute global Histogram of Optical Flow Orientation (HOFO) on a per-frame basis and model normal behaviour using two separate methods, one-class SVM learning, and Kernel Principal Component Analysis embedding. The authors show that the two approaches are effective at modelling normal behaviour when used in conjunction with their proposed descriptor; the one-class SVM offered slightly greater performance. Chen *et al.* [6] extracted a notion of crowd acceleration and stated that rapid changes in acceleration can be used to identify a crowd displaying normal attributed from a crowd currently undergoing a situation of panic. Recent work by Biswas *et al.* [4] express the problem of abnormal behaviour analysis as identifying sparse, or rare behaviours. Each frame is represented as a matrix of features and matrix decomposition is applied to separate the matrix components into two groups, low-rank and sparse components; the latter of which is considered anomalous.

The Violent Flows (ViF) method was proposed by Hassner et al. [17] to identify violent crowds in densely populated areas using changes in optical flow magnitude. Gao et al. [11] state that the ViF descriptor does not capture potentially important changes in orientation and therefore introduced a variant of the ViF descriptor (OViF) that utilises both orientation and magnitude of optical flow. It was shown that ViF offers greater classification ability on crowded data when compared to OViF, but when combined they achieve greater accuracy. Riberio et al. [30] introduce the Rotation Invariant Motion Coherence RIMOC feature that is based on the eigen-values of second order statistics extracted from a Histogram of Oriented Flows. A multi-scale structure is used to model spatio-temporal configurations of features. The authors assume that violent behaviour is unstructured and aim to distinguish violence from non-violence by analysing the likelihood of a feature belonging to a model of normality.

An alternative approach for modelling motion involves tracking features to obtain motion trajectories. Zhou et al. [40] train a Multi-Observation Hidden Markov Model (MOHMM) using motion trajectories extracted from footage of normal behaviour using a Kanade-Lucas-Tomasi Feature Tracker (KLT). The probability of the trained model producing a given observation is computed, if the probability falls beyond a threshold then the observation is considered abnormal. Marsden et al. [24] utilize a KLT tracker to extract motion trajectories and compute holistic measures of crowd collectivness, conflict, and density. These measures form a feature vector that describes the dynamics of a scene. The approach described by Zhou et al. is applicable to many domains of crowd abnormality as it does not assume any specific measurement of crowd motion, whereas the holistic approach by Marsden et al. is useful on data where the measures documented are known to exist. Although tracking has shown to perform well at describing crowd behaviour, Yang et al. [39] highlight the difficulty in tracking when analysing scenes with changing illumination, a property common of naturalistic environments.

One issue with computing measures of motion of a crowd with great accuracy stems from self-occlusion. Occlusion reduces inter-frame correspondences resulting in poor flow approximation or tracking when using traditional optical flow based methods. Ali et al. [1] utilize a particle advection process to extract motion trajectories that describe the underlying flow of a crowd, the process of particle advection smooths trajectories and introduces robustness to minor occlusion. The motion trajectories are then used to generate a Finite Time Lyapunov Exponent (FTLE) field, and subsequently segment a crowd based on motion. Using the particle advection process, Mehran et al. [26] formulated a Social Force Model (SFM) that described the interaction force between pedestrians, a concept defined as a function of desired movement and actual movement. Normal interaction force samples were used to train a Latent Dirichlet Allocation (LDA) model of normality. Raghavendra et al. [28] extended the work by Mehran et al. by using a particle swarm optimization process to minimize the interaction force so to model the most typical crowd behaviours. Yang et al. [38] also built upon the work by Mehran et al. by integrating crowd density that when combined with the SFM created a measure of pressure between pedestrians.

Rao et al. [29] highlight the usefulness of GLCM based crowd description and used GLCM measures to describe the spatial composition of tracked objects in a crowd. Early research by Marana [22] formulated the crowd density estimation problem as a global measure of visual texture. Marana showed that sparse and dense crowds hold notably different textural compositions. Researchers [3, 8, 22, 33] have used the Grey Level Co-Occurrence Matrix (GLCM) approach to crowd description and have shown that Haralick's GLCM features can be used to successfully determine the density of a crowd. The implication being that texture can provide a meaningful description of the visual appearance crowds.

## 3 Proposed Method Overview

Our proposed method builds upon Haralick texture features [15] which describe visual texture using statistics derived from co-occurring grey level intensities. We compute Haralick features for each frame in a sequence and describe how these features evolve over time using simple summary measures to provide a succinct and powerful descriptor of crowd dynamics that yields fast compute time and robustness to change over time. Haralick texture features are extracted from a grey level co-occurrence matrix (GLCM). A GLCM is generated by counting the co-occurring grey level intensity values found in an image given a linear spatial relationship between two pixels. The spatial relationship is defined by a parameter pair $(\theta, d)$ where $\theta$ is the orientation and $d$ is the distance between two pixels. It is common to define a set of parameter pairs $(\theta, d)$ and to then combine GLCM matrices, this is typically used to provide rotational invariance by using a set of orientation parameters, typically in 8 orientations, spaced $\pi/4$ radians apart. The number of grey level values $N_g$ represents the number of unique intensity values present in an image. It is common to scale an image from $[0, 255]$ to $[0, N_g]$ before computing a GLCM, where $N_g$ is a defined number of gray-levels [15].

We compute the following features as defined by Haralick [15]: *Energy*, *Contrast*, *Homogeneity*, *Correlation* and *Dissimilarity*. The variable $P(i, j)$ expressed in Equa-tions (1-5) refers to the value at the $(i,j)^{th}$ position in a gray level co-occurrence matrix.

$$\text{Angular Second Moment} = \sum_{i,j=0}^{N_g-1} P_{i,j}^2 \quad (1)$$

$$\text{Contrast} = \frac{\sum_{i,j=0}^{N_g-1} P_{i,j}(i-j)^2}{(N_g-1)^2} \quad (2)$$

$$\text{Homogeneity} = \sum_{i,j=0}^{N_g-1} \frac{P_{i,j}}{1+(i-j)^2} \quad (3)$$

$$\text{Correlation} = \frac{\left[\sum_{i,j=0}^{N_g-1} P_{i,j} \left[\frac{(i-\mu_i)(j-\mu_j)}{\sqrt{(\sigma_i^2)(\sigma_j^2)}}\right]\right]+1}{2} \quad (4)$$

$$\text{Dissimilarity} = \frac{\sum_{i,j=0}^{N_g-1} P_{i,j}|i-j|}{N_g-1} \quad (5)$$

The equations for contrast (Equation 2), correlation (Equation 4) and dissimilarity (Equation 5) are altered such that the returned value is bounded between $[0,1]$. Given a series of images expressing appearance over time, we compute the aforementioned texture features from the resulting time ordered sequences of texture measure over time, each referred to as $x$. We calculate the statistical summary of each sequence to encode the underlying crowd behaviour. Each sequence $x$ is represented as a 4 length feature vector composed of measures of mean, standard deviation, skewness (Equation 6) and inter-frame uniformity (IFU, Equation 7). Skewness indicates the asymmetry found in a distribution and can be used to deduce whether a distribution is showing a general increase or decrease in value over time. Inter-frame Uniformity (IFU) as expressed in Equation 7 is a measure of adjacent sample similarity within time ordered data. It is expressed as the scaled $L_2$ norm (Equation 7) of the sequence $y$ where sequence $y$ is formed by taking the absolute difference between adjacent samples within sequence $x$, $y_t = |x_t - x_{t+1}|$. Sequence $y$ is normalized by its sum before being input into Equation 7. IFU returns values within the range $[0,1]$ where 0 and 1 represent non-uniform, and uniform change over time respectively. This particular measure of uniformity was designed to be sensitive to sudden change in value over time and is therefore intended to be suited towards highlighting more abrupt changes in time-ordered data.

$$\text{Skewness} = \frac{E(x-\mu)^3}{\sigma^3} \quad (6)$$

$$\text{IFU} = \frac{|y|_2\sqrt{(T-1)}-1}{\sqrt{(T-1)}-1} \quad (7)$$

It was observed that different spatial regions in frame depicted different behaviour, therefore we spatially split each video into $M \times N$ non-overlapping sub-regions and apply the aforementioned method to each. Each cell is represented by twenty values that describe the appearance, and change in appearance over time. We generate twenty histograms, one for each feature, using values taken from each cell within the $M \times N$ grid. Through empirical analysis we found that using logarithmically distributed histogram bins within the range of $[0,1]$ provided the best performance. Histograms representing skewness are bounded between $[-1.4, 1.4]$ and the bins are logarithmically, and symmetrically distributed around zero such that bin spaces are closer at values closer to zero.

In the case of surveillance footage, failure to remove background information may lead to the description of landmarks as opposed to crowd dynamics. Two GLCMs generated adjacent in time will have a very similar composition as static objects will introduce the same information in both matrices. To remove static information that typically corresponds to background information, we subtract adjacent GLCMS, $M_t - M_{(t-1)}$ and threshold all values less than 0 where $t$ represents the frame being analysed. This approach comes at a near negligible computational cost and offers robustness to minor translational camera motion due to the spatially unconstrained nature co-occurrence matrices.

## 4 Tested Data

The goal of our research is to provide a computational method that can aid CCTV operatives at detecting scenes that exhibit either violent or abnormal behaviour. In this section we outline four datasets and describe their attributes. As a brief description of purpose, the CF-Violence and Violent Flows datasets are used to evaluate the ability of the proposed method at detecting violent behaviour, whereas the UMN and UCF Crowd Abnormality datasets are used to evaluate the ability of analysing more general crowd behaviour.

### 4.1 CF-Violence Dataset

We obtained real-life surveillance footage from a local police force that showed either violent or non-violent behaviour within city centre locations. Experiments performed on this data will provide a realistic understanding of the applicability of each tested method in a real-world scenario.

We obtained 13 samples of violent behaviour and 63 samples of general behaviour. The violent scenes can be separated into two distinct classes of *high* and *low* based on the participant population. Only 4 of the 13 samples can be considered to have a high number of participants. Video resolutions range between $320 \times 240$ and $640 \times 480$ and all videos were recorded at a de-interlaced frame-rate of six frames per second. Surveillance cameras are typically placed at a high altitude in order to maximize viewshed. The elevated height increases a cameras exposure to high wind speeds which causes the camera to shake and capture spatially unstable footage; this can cause issues when trying to identify corresponding features between adjacent frames. All real-world footage is stabilized before subsequent analysis using the state-of-the-art method stabilization method proposed by Grundmann et al. [14].

### 4.2 Violent Flows Dataset

The Violent Flows dataset [17] was created for the sole purpose of evaluating crowd violence classification methods; it is a relatively new dataset and is not widely tested. There are 123 instances of both violent and non-violent data samples available from footage uploaded to video media hosting websites. The violent footage contains many samples that are visually similar to those found in real-world data and it is therefore suitable for evaluating the violence classification properties of our proposed method.

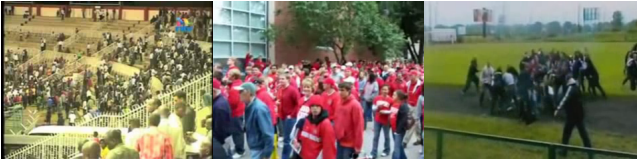

**Fig. 1** Examples frames taken from the Violent Flows crowd violence dataset

### 4.3 UCF Web Crowd Abnormality Dataset

The UCF web crowd abnormality dataset consists of 20 videos depicting either normal or abnormal crowd behaviour [26]. Abnormal data is classified as *panic*, *clash* or *fight* scenarios. Normal samples can be described as showing either crowds walking in an urban environment or pedestrians running in a marathon. The dataset has been obtained from various media hosting websites and of the 20 available sequences, 12 are normal and 8 abnormal. Footage is recorded at 24 frames per second with a resolution of $640 \times 480$. Image examples of the web crowd abnormality dataset can be seen in Figure 2.

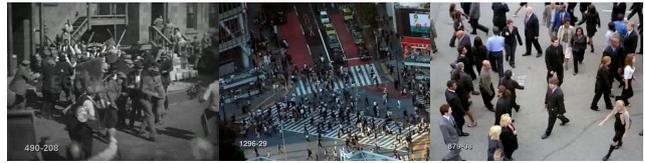

**Fig. 2** Examples frames taken from the Web Crowd Abnormality dataset

### 4.4 UMN Crowd Abnormality Datasets

The UMN unusual crowd activity dataset [26] is a synthetic dataset that depicts sparsely populated areas. Normal crowd activity is observed until a specified point in time where behaviour rapidly evolves into an *escape* scenario where each individual runs out of camera view to simulate panic. The dataset is comprised of 11 separate video samples that start by depicting normal behaviour before changing to abnormal. The panic scenario is filmed in three different locations, one indoors and two outdoors. All footage is recorded at a frame rate of 30 frames per second at a resolution of $640 \times 480$ using a static camera.

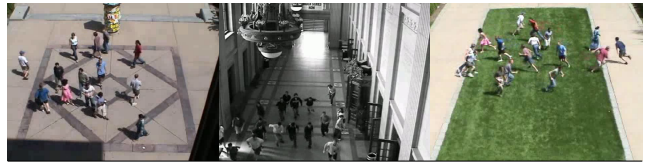

**Fig. 3** Examples frames taken from the UMN Unusual crowd dataset

## 5 Results

A classification label is generated for each video frame in order to provide a continuous activity feed usable in CCTV observation scenarios. This is achieved by classifying a description vector computed using the previous *n* frames in sequence, by default we assign *n* to be equal to the number of frames per second. For the generation of the grey level co-occurrence matrix we assign $N_g = 32$. The parameters $(\theta, d)$ are assigned as $(0, 1)$, see section 5.1 for an explanation regarding the choice of these parameters. *M* and *N*, which specify frame sub-division used to encode spatial information are assigned the value of 4, section 5.2 discusses the effects of using different values for *M* and *N*.

All experiments were conducted using *k*-fold cross validation where data is split into *k* partitions with $k-1$ partitions being used for training a random forest classifier [5]. The remaining partition is used for testing; the random forest is composed of 50 trees. The parameter *k* is assigned a value of 5, 5, 2, and 2 when processing the CF-Violence, Violent Flows, UMN and Web Abnormality datasets respectively.

We perform each experiment 10 times and report the average result to reduce any variability introduced by random sampling during cross validation. As stated previously, we extract features such that each frame in a sequence is represented by a single vector, we do not allow features extracted from a single source video to be placed in both training and testing partitions at the same time, as features extracted from any single video are likely to belong to the same distribution and may lead to over-fitting. We present results using receiver operating characteristic (ROC) curves, a common way to summarise these curves is to report area under the curve. Area under ROC dictates the discrimination performance between binary classes, a value of 1 indicates perfect discrimination. Our method was implemented using Python and the Skimage library, and we perform all experiments using an Intel i7-4790 at 3.6GHz processor. Given a temporal window size $n$ of 24, and a resolution of $640 \times 480$, our method operates at 76.92 frames per second, or 0.013 seconds per frame.

Decomposing the importance of temporal features we find that the measure of Intra-frame Uniformity is highly descriptive (Figure 4) when applied to the two datasets whose *abnormal* class contains only violent samples, these being the CF-Violence and Violent Flows datasets. Looking at the average IFU values returned by these datasets, we find that the appearance of violent scenes within the Violent Flows dataset change in a less uniform manner over time (Table 1). We also observe that appearance of scenes in the CF-Violence dataset, as represented by ASM and Homogeneity measures, exhibit the same property. Given this observation, we formulated an additional experiment to deduce whether or not a lower IFU is indicative of violent behaviour when compared to normal behaviour. The Web Abnormality dataset contains examples of violence within the *Abnormal* class, we have separated all non-violent abnormal scenes to create two new binary datasets, these are *Violent* or *Normal* (VoN), and *Violent* or *Abnormal* (VoA), the latter differs in that the *Abnormal* class is composed of the Non-Violent abnormal samples from the Web Abnormality dataset. We observe that the IFU measure reported across all appearance features for both VoN and VoA (Table 1), is less for scenes of violence, this suggests that violence has a greater non-linear change in appearance over time when compared to non-violence. Continuing the IFU analysis, we find that scenes of abnormality as displayed in the UMN dataset have a greater IFU than scenes of normality, this highlights that a low IFU is not indicative of all types of abnormal behaviour.

When testing the UMN dataset our proposed approach achieves comparable classification ability to other state-of-the-art methods (Table 3) when using a temporal window size greater than 64 frames in length (Table 9). We find that the rate of classification increases as the temporal window length increases. It is believed that as the panic situ-ation winds down, the key characteristics of panic are less prominent, therefore increasing the temporal window size prolongs the time in which the dominant characteristics remain in the decision making process.

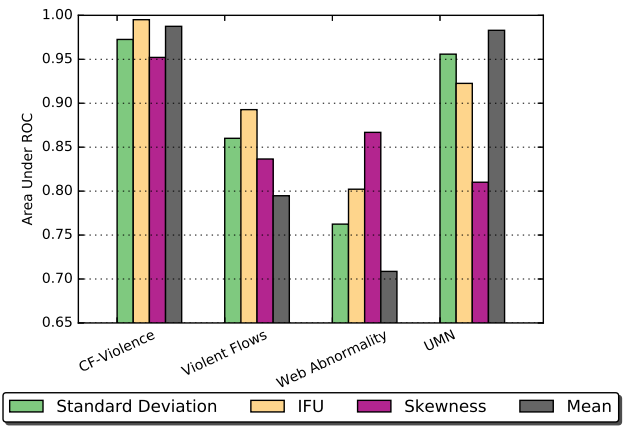

**Fig. 4** Classification performance achieved by each temporal feature type.)

We observe that the measure of mean appearance over time is the a weakest descriptor when applied to the Web Abnormality and Violent Flows datasets (Figure 4), we hypothesise that the appearance of crowds within these datasets vary a lot as the are recorded from different sources, therefore strong visual correspondences in appearance across samples is unlikely. Both the UMN and CF-Violence dataset use fixed cameras which record different crowd behaviours within the same environment. Given that the environment can guide the flow/behaviour of a crowd, then typical crowd compositions emerge during scenes of normality, in which case the mean appearance offers high classification ability as inter-sample visual similarity is more likely to occur between samples that depict normality.

We conducted evaluation on the Violent Flows dataset using the approach outlined in the seminal paper by Hassner *et al.* [17]. We demonstrate that our method offers comparable performance with existing methods (Table 5). We also found that the only alternative method to report a greater classification accuracy than our approach does not operate in real-time [13], unlike our own approach which does boast real-time performance.

### 5.1 Pixel pair relationship

As stated in section 3, a grey level co-occurrence matrix is generated by counting pixel pair occurrences given a relationship defined by parameters $(\theta, d)$. We want to investigate the effects of $(\theta, d)$ to determine whether or not a common value exists that offers good performance across different data types. We evaluate the effects of these parameters by performing multiple experiments with a range of

Detecting Violent and Abnormal Crowd activity using Temporal Texture Analysis 7

| IFU | Dissimilarity | Correlation | Homogeneity | ASM | Contrast |
|---|---|---|---|---|---|
| CF-Violence | 0.000781146 | 0.0161029 | -0.00154672 | -0.0137449 | 0.0105682 |
| Web Abnormality | -0.00434525 | -0.00614079 | -0.00603542 | -0.0140028 | -0.00323114 |
| UMN | 0.00240629 | 0.00131867 | 0.00207164 | 0.00343199 | 0.00591058 |
| Violent Flows | -0.00723042 | -0.00840384 | -0.00728228 | -0.028443 | -0.00750538 |
| VoA | -0.00563025 | -0.00444586 | -0.00483273 | -0.00330466 | -0.00513838 |
| VoN | -0.00784559 | -0.00890479 | -0.00903995 | -0.0160573 | -0.00642569 |

**Table 1** Inter-Frame Uniformity measure difference between scenes of Abnormality and Normality for each texture measure outlined in Section 3. Negative values indicate Normal scenes have a greater value than Abnormal scenes, indicating that scenes of Normal behaviour are temporally more uniform.

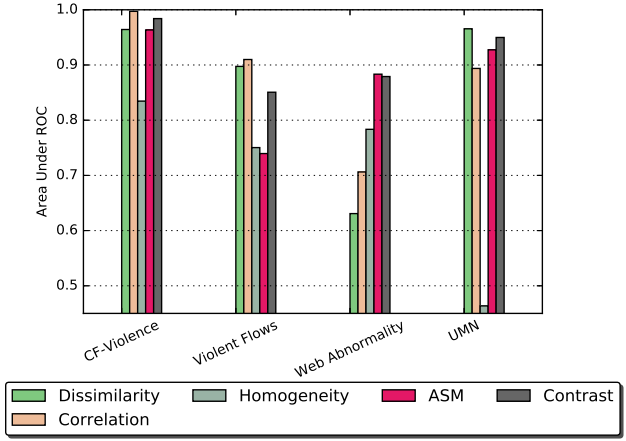

**Fig. 5** Classification performance achieved by each texture feature type.)

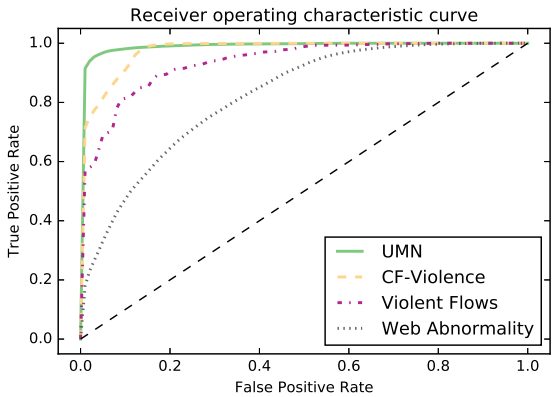

**Fig. 6** ROC curves for each tested dataset

| Method | AUC |
|---|---|
| Proposed | 0.9782 |
| ViF [17] | 0.80 |
| OViF [11] | 0.76 |
| Fast Fight [13] | 0.89 |

**Table 2** CF-Violence classification score.

| Method | AUC |
|---|---|
| Proposed | 0.9956 |
| Optical Flow [26] | 0.84 |
| SF [26] | 0.96 |
| MDT [21] | 0.9965 |
| Chaotic Invariants [36] | 0.99 |
| Biswas [4] | 0.9838 (Average) |

**Table 3** UMN classification performance scores including state-of-the-art results.

| Method | AUC |
|---|---|
| Proposed | 0.8218 |
| SF [26] | 0.73 |
| Optical Flow [26] | 0.66 |

**Table 4** UCF Web Abnormality Crowd dataset classification performance scores including state-of-the-art results.

| Method | Accuracy (±) | AUC |
|---|---|---|
| Proposed | 86.03 ± 4.25% | 0.9403 |
| Fast Fight [13] | 69.40 ± 5.0 % | 0.7500 |
| ViF [17] | 81.30 ± 0.21 % | 0.8500 |
| OViF [11] | 76.80 ± 3.90 % | 0.8047 |
| Holistic Features [24] | 85.53 ± 0.17 % | – |
| MoSIFT [37] | 83.42 ± 8.03 % | 0.8751 |
| MoSIFT (KDE / Sparse Coding) [37] | 89.05 ± 3.26 % | 0.9357 |

**Table 5** Violent Flows dataset classification performance scores including state-of-the-art results.

of 8 orientation values spaced $\pi/4$ radians apart, the second set contains 4 orientation values spaced $\pi/2$ radians apart. The final orientation set contains a single value of 0. The five distance values are a sequence of integers that double in size starting from 1. The results of this experiment are shown in Figure 7. Across all tested datasets we observe a common trend in that orientation has no significant impact on descriptive ability. We conducted a second experiment in which we randomly rotated each cell by either 0, 90, 180 or 360 degrees, even in this case we found no significant difference in classification performance when using each of the three orientation compositions.

When analysing the results from the Web Abnormality, Violent Flows and CF-Violence datasets, we observe a negative correlation between classification performance represented by area under ROC and parameter $d$; the UMN datasets shows a less uniform relationship. We hypothesis this pattern occurs due to the distance between interacting entities within each video, for instance, the crowds found in the CF-Violence and Violent Flows dataset can be described as densely populated and as a result the pedestrians have a close proximity to one another. In contrast to this, the UMN dataset is sparsely populated and the distance between moving entities is much larger. In the close proximity scenario, a small distance value is better suited towards identifying meaningful relationships between two interacting entities, conversely, in a sparse scenario a small value for $d$ may not be great enough to relate two distant entities, in which case a greater value of $d$ should be chosen. Ultimately we find that the parameter pair $(0, 1)$ provides the best performance across all datasets.

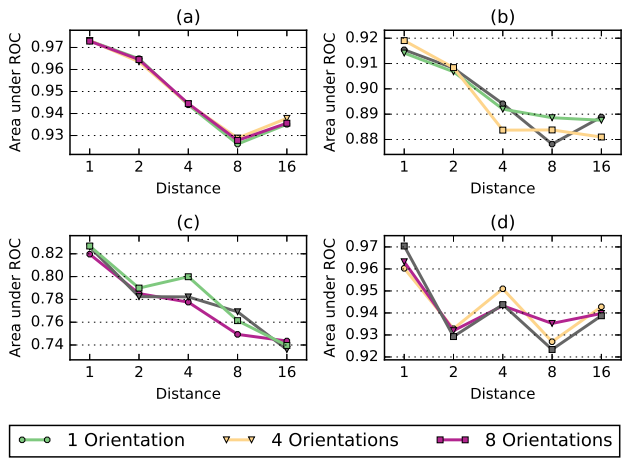

**Fig. 7** Graphs show the effects of pixel pair relationship parameters: a) CF-Violence, b) Violent Flows, c) UCF Web Abnormality, d) UMN Panic

5.2 Frame Split

In section 3 we suggest splitting a scene into sub-regions referred to as cells, we then extract a description for each cell and aggregate them to form a global descriptor of a scene. This is performed as a scene may consist of different local behaviours that cannot be strongly represented when processing the entire scene as a single cell ($M = N = 1$). The Violent Flows dataset sees maximal classification score when $N = M = 2$. When the scene decomposition becomes too fine $M = N > 8$ the classification performance drops, a similar trend occurs when analysing the Web Abnormality dataset. We hypothesise that a larger cell size is more suitable for describing the global characteristics of behaviour in a dense crowd. Using a fine grid results in the description of small components such as individuals, in which case the characteristics encoding the effect of an individual on a crowd is less explicitly encoded as the local cell aggregation process used to form the global descriptor discards spatial locality of behaviour, and therefore any relationship between an action and its associated reaction is also discarded. Larger cell sizes will encapsulate multiple people and therefore describe the interaction, both action and reaction. The CF-Violence dataset does not follow the same trend, it shows a positive correlation between grid size and classification score, suggesting that in this case, small individual components of a scene are sufficient enough to describe violent behaviour. This is reasonable when you consider the contrast between normal and abnormal behaviour during the NTE, for instance, violent acts such as kicking or punching are vastly dissimilar to the typical types of normal behaviour, therefore smaller cells that encapsulate individual actions are still capable of encoding abnormal behaviour as its the action, not the interaction that matters. In contrast, the difference between the individual actions of people within the Violent Flows dataset during violent and non-violent scenes is less clear, and so the interactions are important, and as stated prior, encoding interactions require larger observation windows.

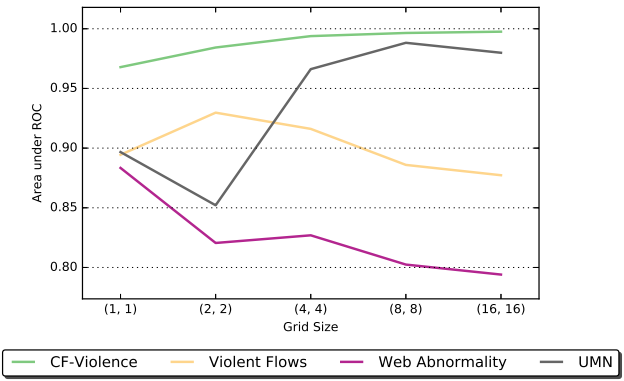

**Fig. 8** The effect on accuracy of using different values for $M$ and $N$ where $M = N$

5.3 Window Length

In this subsection we will investigate the effects of parameter $n$ to determine if the description of crowd behaviour is best formulated using either short or long term temporal dynamics. We perform classification using descriptors formed using the following values of $n$: 6, 12, 24, 32, 64 and 128. We find that all datasets favour larger window sizes (Figure 9), suggesting that the distinction between scenes of normality and abnormality is made more clear over long periods. Although each dataset has its preferential window length, it is important to note that short window sizes still offer reasonably good performance across all datasets. When transitioning from normal to abnormal behaviour, the amount of time required for the majority of the feature vector to be composed of information from abnormal behaviour will be greater the larger the observation window. Assuming that class transitions are not represented by the descriptor, the worst case for classifying abnormal behaviour will see a delay of at most *n* frames. Therefore shorter observation windows are more appropriate for use in a real-time system as it will allows for more instantaneous updates regarding the dynamics of the scene.

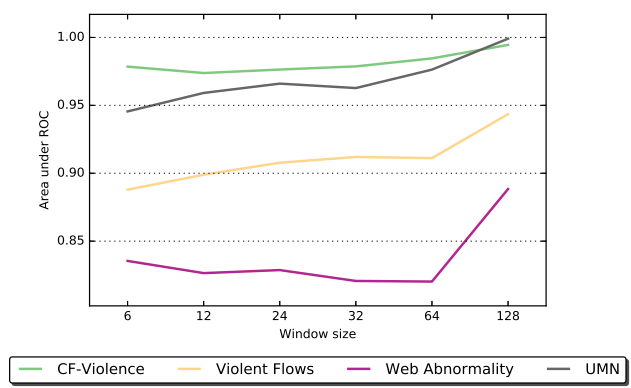

**Fig. 9** The effect on accuracy of using different window sizes *n*

5.4 Data Quality Analysis

As discussed in the introduction (Section 1), the quality of CCTV recorded footage can often be poor. It is useful to investigate the effects of image quality on the ability of computer based methods to operate, as this information can be used to determine, and resolve weaknesses in a system. We test for correlation between measures of image quality and a binary value indicating whether or not a sample was correctly classified. Given that we remove background elements when forming our descriptor we perform background subtraction before computing image quality measures to avoid drawing spurious correlations between our description of the crowd and quality of the background. To remove the background we perform frame differencing and apply a threshold of 0.04 to remove all static regions. Using the point biserial correlation measure we find that in most cases, each measure of image quality displays no significant correlation ($p > 0.05$, Table 6). When analysing the Web Abnormality data we expected to find a negative correlation between noise and prediction, however we observe the opposite. We hypothesise that the measure of noise is incorrectly measuring the features of a crowd as noise given that certain crowds can appear visually noisy. In addition to this we observe that Image Complexity and Contrast Factor are positively correlated suggesting that strong structure is a key requirement for our proposed method to effectively analyse the Web Abnormality dataset.

## 6 Conclusion and Future Work

In this paper we have utilised GLCM texture features that are typically used in crowd density estimation, and applied temporal encoding to create an effective method that describes crowd dynamics. We have demonstrated that the proposed method is highly effective at discriminating between scenes of normal and abnormal behaviour, and that our approach operates in real-time. We have highlighted that violent behaviour typically holds a less uniform rate of change over time when compared to other types of typical crowd behaviour, further analysis must be conducted to identify whether or not this property exists given alternative measurements of crowd behaviour. We have provided an in-depth evaluation of the parameter effects present in our proposed method that should provide insight into the most suitable choice of parameter values, however, further research can be conducted to determine a method of adaptively choosing the optimal parameters given some data. We would also like to apply computer vision based violence detection systems in the real-world to evaluate their ability at reducing the impact of violence related injuries by assisting CCTV observation personnel.

**Table 6** Correlation between each measure of quality and the binary label indicating whether or not a sample was correctly classified.

| Measure | Web Abnormality | | UMN | | Violent Flows | | CF-Violence | |
|---|---|---|---|---|---|---|---|---|
| | r | p | r | p | r | p | r | p |
| Sharpness [2] | 0.038748 | 0.119578 | -0.0829886 | 0.00614006 | -0.0144083 | 0.432805 | 0.0106051 | 0.576142 |
| Noise [18] | 0.155387 | 3.43226e-10 | -0.064551 | 0.0331752 | -0.00693547 | 0.705759 | 0.00140241 | 0.941071 |
| Contrast Factor [25] | 0.113564 | 4.76052e-06 | 0.0311274 | 0.304762 | -0.002687 | 0.883704 | 0.0159766 | 0.399675 |
| Colourfulness [16] | 0.0612761 | 0.0137814 | 0.0243059 | 0.422963 | -0.00202095 | 0.912396 | 0.000616108 | 0.974092 |
| Image Complexity [20] | 0.145287 | 4.48378e-09 | -0.0778508 | 0.010169 | -0.00640437 | 0.727356 | 0.00216641 | 0.909083 |